\documentclass[letterpaper, 10 pt, conference]{ieeeconf}  

\IEEEoverridecommandlockouts                              

\usepackage{graphics} 
\usepackage{graphicx}
\usepackage[space]{grffile}
\usepackage[export]{adjustbox}
\usepackage{epsfig} 
\usepackage{amsmath} 
\usepackage{amssymb}  
\usepackage{booktabs}

\usepackage[abs]{overpic}
\usepackage[usenames, dvipsnames]{color}
\usepackage{soul}

\graphicspath{{fig-arxiv/}}

\title{\LARGE \bf
Material Recognition CNNs and Hierarchical Planning for Biped Robot Locomotion on Slippery Terrain
}

\author{Martim Brand\~{a}o, Yukitoshi Minami Shiguematsu, Kenji Hashimoto and Atsuo Takanishi
\thanks{*This work was supported by JSPS KAKENHI Grant Number 15J06497 and ImPACT TRC Program of Council for Science, Technology and Innovation (Cabinet Office, Government of Japan).}
\thanks{M. Brand\~{a}o and Y. M. Shiguematsu are with the Graduate School of Advanced Science and Engineering, Waseda University.}
\thanks{K. Hashimoto is with the Waseda Institute for Advanced Study, and is a researcher at the Humanoid Robotics Institute (HRI).}
\thanks{A. Takanishi is with the Department of Modern Mechanical Engineering, Waseda University; and the director of the Humanoid Robotics Institute (HRI), Waseda University.}%
}

\begin{document}

\maketitle
\thispagestyle{empty}
\pagestyle{empty}

\begin{abstract}

In this paper we tackle the problem of visually predicting surface friction
for environments with diverse surfaces, and integrating this knowledge 
into biped robot locomotion planning.
The problem is essential for autonomous robot locomotion since diverse surfaces 
with varying friction abound in the real world, from wood to ceramic tiles, grass or ice, 
which may cause difficulties or huge energy costs for robot locomotion if not considered.
We propose to estimate friction and its uncertainty from visual estimation of 
material classes using convolutional neural networks, together with probability 
distribution functions of friction associated with each material. 
We then robustly integrate the friction predictions into a hierarchical (footstep and full-body)
planning  method using chance constraints, and optimize the same trajectory costs
at both levels of the planning method for consistency.
Our solution achieves fully autonomous perception and locomotion on slippery terrain,
which considers not only friction and its uncertainty, but also collision, stability and 
trajectory cost.
We show promising friction prediction results in real pictures of outdoor scenarios, 
and planning experiments on a real robot facing surfaces with different friction.

\end{abstract}

\section{INTRODUCTION}
\label{sec:introduction}

Legged and humanoid robot locomotion planning is an important problem for disaster response and service robots.
One of the difficulties of this problem is the complexity of general environments
and the need to consider several factors such as collision, energy consumption, surface geometry and friction.
In this paper we deal with the specific problem of humanoid robot locomotion when environment
friction is considered.
Our claim is that friction and its uncertainty can be estimated from vision and robustly 
integrated into algorithms for motion planning with contact.
We argue that even if the precise coefficient of friction cannot be predicted from vision before touching a surface,
priors and accumulated experience associated with surface material or condition (think coefficient of friction tables) 
can provide a probability distribution of friction. 
Motion planning with contact can also become prohibitively expensive once multiple factors are considered, 
such as locomotion cost, collision and friction. In this paper we propose a hierarchical approach to the problem,
where a footstep planner optimizes the same cost function as a full-body motion planner by use of an 
oracle, and considers collision and friction by using simple bounding box collision checks and an 
``extended footstep planning'' \cite{Brandao2016tro} approach.

The contributions of this paper are the following:
\begin{itemize}
\item We propose a solution to the friction from vision problem using a state-of-the-art deep Convolutional Neural Network (CNN)
architecture to predict broad material classes from images, together with known (or learned) distributions of material friction;
\item We propose a hierarchical planning architecture for biped robots that optimizes the same objective at both levels, 
and deals with friction, stability, collision and cost to produce full-body trajectories;
\item We show empirical friction prediction results, as well as planning experiments which show the usefulness
and applicability of the approach in complex environments with varying friction.
\end{itemize}

\section{RELATED WORK}
\label{sec:related}

Recent full-body motion planning \cite{Kuindersma2016,Hauser2008} and control \cite{Kuindersma2016,Herzog2016,Feng2013} algorithms for legged robots
have started to consider friction by using friction cones in optimization problems.
Such methods rely on the fundamental assumption that the coefficient of friction can be 
predicted in advance. While in itself a challenging problem, partial evidence from human 
visual perception motivate such an approach to the problem. 
For example, humans are known to use visual cues to estimate friction, related to surface 
texture \cite{Lesch2008}, shine \cite{Joh2006} and detection of materials or contaminants (e.g. water) \cite{Li2004}.
Furthermore, in the human gait literature there is evidence that humans use accumulated 
previous experience to predict friction and adapt walking style before touching slippery 
ground \cite{Cappellini2010}.
In this paper, such ``accumulated previous experience'' is implemented as probability 
distributions of friction associated with material classes. The term ``material'' is used 
in a broad sense to refer to visually classifiable classes related to material, condition 
and context (e.g. ``dry metal'', ``wet asphalt road'').
The use of material classes for prediction here is motivated by a recent study \cite{Brandao2016friction} which identifies material as one of the most predictive features of both coefficient of friction values and human judgements of friction.

Friction estimation work related to this paper includes that of Angelova et. al \cite{Angelova2006}, 
which predicts the percentage of slip (i.e. lack of locomotion progress) of a rover from terrain 
classification and slope. Predictions are made based on non-linear regression of data gathered on 
a learning stage. Compared to \cite{Angelova2006}, we estimate the coefficients of friction 
instead of slip, we use open segmentation datasets \cite{Bell2013,Mottaghi2014} to train 
material classification, and we decouple the problem from the physical robot. 
Importantly, our approach allows for sharing material friction data among different 
robots as long as they have similar foot soles. 

Several approaches exist to the friction-constrained motion planning problem for legged robots.
One approach is the non-hierarchical, full-scale trajectory optimization 
formulation with implicit contact constraints of \cite{Posa2013,Mordatch2012}. 
While technically elegant and showing promising results, these can still be
computationally expensive for online planning.
In order to make the problem tractable, full-body motion can be planned after 
contact (or footstep) planning \cite{Deits2014,Escande2013}, in what is
called the \textit{contact before motion} approach.
One common issue with such methods is that contact planners do not take the same 
friction or trajectory criteria into account as the subsequent full-body planners.
One exception to this lack of consistency between planning levels is the 
\textit{extended footstep planning} work of \cite{Brandao2016tro}, 
in which learned models are used at the footstep planning level that predict
full-body feasibility.
The approach can also account for friction constraints by using timing variables 
and learned slippage models at the footstep planning level. 
Still, in \cite{Brandao2016tro} the costs optimized at the footstep planning level are 
not further optimized at the full-body level. In this paper we improve the method by 
using an oracle at the footstep planning level which predicts the costs obtained by a 
full-body trajectory optimizer, thus increasing consistency across planning levels.

\section{FRICTION FROM VISION}
\label{sec:vision}

In this paper we propose to estimate friction of surfaces from visual input by
classifying surface material at each image pixel and assuming known (or learned) probability 
distributions of friction for each material.
For convenience we will use the term ``friction of a material'' to refer to the coefficient of friction
between the robot foot sole and a second surface of a given material.

We consider a pixel-wise labelling algorithm that, given an input image $I$ with $n$ pixels, 
provides a probability distribution $P(X | \theta, I)$, where $X=\{x_1,...,x_n\}$ are the 
pixel labels and $\theta$ are internal parameters of the algorithm.
Each pixel can take one of $m$ possible labels, such that $x_k \in \mathcal{L} = \{ l_1,...,l_m \}$.
Furthermore, let each label be a material associated with a probability distribution function (p.d.f.) 
of a coefficient of friction $p(\mu | l_i)$.
Then at pixel $k$, the conditional p.d.f. of $\mu$ is
\begin{equation}
p(\mu | \theta, I) = \sum\limits_{i=1}^m p(\mu | l_i) P(x_k=l_i | \theta, I) .
\label{eq:friction}
\end{equation}

For the results shown in this paper we estimated the friction distributions $p(\mu | l_i)$ experimentally, 
by measuring maximum friction force of the robot foot on several surfaces for each material.
We describe the procedure in more detail in Section \ref{sec:results}.

We use a deep convolutional neural network (CNN) to obtain pixel-wise material predictions
$P(x_p=l_i | \theta, I)$.
In particular we use the encoder-decoder architecture of \cite{Badrinarayanan2015}, which
achieves good results in image segmentation applications and is characterized by a low number of
parameters. Its low number of parameters leads to fast inference, which is crucial for robotics.
The architecture consists of an encoder network of 13 convolutional layers as in VGG16 \cite{Simonyan2014}, 
followed by a decoder network of 13 layers and a final softmax layer.
The output of the last layer of the network (a softmax classifier) is at each pixel a vector of 
probabilities for each class, that is, the probabilities $P(x_p=l_i | \theta, I)$ used in equation (\ref{eq:friction}).

\section{HIERARCHICAL PLANNING}
\label{sec:planning}

\begin{figure}
\centering
\includegraphics[width=1\linewidth]{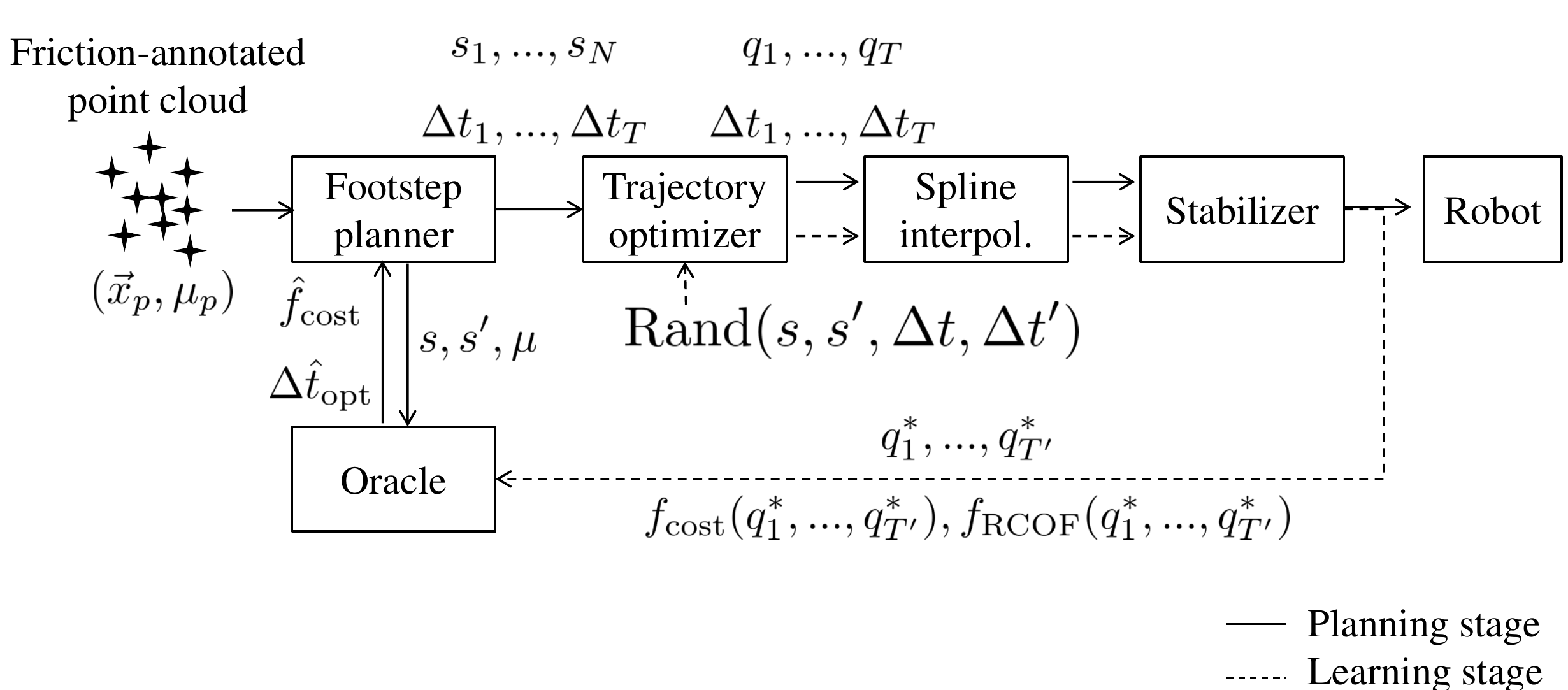}
\caption{Our hierarchical planning architecture, which uses trajectory optimization to minimize
	a cost function $f_\mathrm{cost}$, as well as oracle costs to plan footstep placement and 
	timing that will have low predicted $f_\mathrm{cost}$.}
\label{fig:flowchart}
\end{figure}

In this paper we plan full-body robot motion using a \textit{contact before motion} approach. 
A footstep planner first searches a stance graph using transition costs provided by an oracle. 
The stances are then used as constraints in a full-body trajectory optimizer that considers
full-body trajectory costs, collisions, joint limits and static stability.
The obtained trajectory is finally interpolated and locally adapted for dynamic stability using a ZMP-based method.
The oracle basically takes each stance transition and predicts the costs obtained at the end of the
whole planning pipeline. This leads to footstep plans which optimize the same criteria as the 
full-body planner.
See Figure \ref{fig:flowchart} for a visual representation of the architecture.

\subsection{Extended footstep planning with an oracle}
\label{sec:planningFootstep}

The footstep planner searches a graph of stances to find a feasible path between the start and goal stance.
Each node in the graph is a stance $s$, which is defined by a set of contacts with the environment.
A contact is a tuple (link, position, rotation).
A neighbor stance $s'$ either adds or removes a contact with respect to $s$.
In this paper we deal with biped walking only, and hence stances simply transition 
from double-support to left-foot-contact, to double-support, to right-foot-contact, back to double-support, etc.
The advantage of this representation instead of, for example, double-support stances only, is that the swept-volume between consecutive stances can be used by the optimizer to guide a swing leg out of collision. 
Such an approach is also used by other works focusing on collision detection \cite{Perrin2012}.

In this paper we use the \textit{extended footstep planning} framework of \cite{Brandao2016tro}. 
The footstep planner is ``extended'' because extra parameters associated with stance transitions (e.g. step timing)
are computed from the transition itself by a function learned offline. 
Here we call this function an oracle because it predicts the costs that will be obtained
by a subsequent full-body trajectory optimizer.

We now briefly describe the footstep planning algorithm.
We first discretize the search space by constraining contact positions to a point cloud, and rotations by aligning 
contact normals with the environment and constraining the links' yaw orientation to a discrete set of values 
in the global coordinate frame.
Then we use an A* variant, ARA* \cite{Likhachev2003}, to search the stance graph based on oracle costs.
At each state (i.e. stance) of the graph $s$, a contact is either added or removed to 
generate successor stances.
Contact removal generates one new stance.
On the other hand, adding a new contact consists of doing a range search of points
in a radius around the foot in contact. For each of those points, footsteps are placed
at all yaw angles and checked for feasibility. 
We implement feasibility as empirical stance distance limits, as well as foot-foot and 
COM-environment collision checking using bounding boxes for the feet and trunk. 
The feasible stances are added as successors of $s$.

To find an optimal path to the goal state, A* search requires a state transition cost function 
$c(s,s')$ and a heuristic cost-to-go function $h(s)$.
Similarly to \cite{Brandao2016tro}, we define the cost as
\begin{equation}
\begin{split}
c(s,s') = & \underset{p}{\text{min}} \ \hat{f}_\mathrm{cost}(s,s',p) \\
& \text{subject to} \\
& P(\hat{f}_\mathrm{RCOF}(s,s',p) < \mu^{(k)}) \geq \eta \\
& k=1,...,K ,
\label{eq:astarCost}
\end{split}
\end{equation}
where $k$ is an index of the contacts of $s$ and $s'$, $\mu^{(k)}$ is the friction at these contacts,
and $p$ are state transition parameters.
Since in this paper we use two full-body posture waypoints per stance at the
trajectory optimization level (Section \ref{sec:planning}), we set transition parameters $p=(\Delta t, \Delta t')$.
These are the time spent from the second waypoint of $s$ until the first waypoint of $s'$,
and the time spent from the first waypoint of $s'$ to the second, respectively.
The main difference in (\ref{eq:astarCost}) with respect to \cite{Brandao2016tro} is that 
we consider uncertainty in the coefficient of friction variable by using chance constraints.
RCOF stands for required coefficient of friction and corresponds to the maximum tangential-to-normal 
force ratio exerted over the whole trajectory \cite{Brandao2016tro}. 
Therefore, the constraints $P(\hat{f}_\mathrm{RCOF}(s,s',p) < \mu^{(k)}) \geq \eta$ in (\ref{eq:astarCost}) 
implement robust Coulomb friction at each contact by forcing the inequalities to hold with at 
least probability $\eta$.
The constraints can also be rewritten using the cumulative distribution function of (\ref{eq:friction})
denoted by $F_{\mu^{(k)}|\theta,I}$,
\begin{equation}
F_{\mu^{(k)}|\theta,I}(\hat{f}_\mathrm{RCOF}(s,s',p)) \leq 1-\eta .
\end{equation}
Since each $\mu^{(k)}$ is one-dimensional then $F$ can be inverted and the constraints
rewritten in deterministic form
\begin{equation}
\hat{f}_\mathrm{RCOF}(s,s',p) \leq Q_{1 - \eta}^{(k)} ,
\label{eq:quantile}
\end{equation}
where $Q_{1-\eta}^{(k)}$ is the $(1-\eta)$-quantile of $F_{\mu^{(k)}|\theta,I}$,
which can be computed by an integral of (\ref{eq:friction}) over $\mu$.

Regarding the heuristic cost-to-go function of A* search, as in \cite{Brandao2016tro}, we set it to
\begin{equation}
\begin{split}
h(s) = & d_{xy}(s , s^\mathrm{goal}) . \underset{s, s', p}{\text{min}} 
\frac{ \hat{f}_\mathrm{cost}(s,s',p) }{ d_{xy}(s , s') }  ,
\label{eq:astarHeuristic}
\end{split}
\end{equation}
where the function $d_{xy}(.,.)$ 
computes Euclidean distance on the horizontal plane between two stances 
(i.e. the distance between left feet and right feet summed).
The heuristic (\ref{eq:astarHeuristic}) is a lower bound on the cost-of-transport times distance,
which guarantees that $h(s)$ does not overestimate the total cost to the final stance $s^\mathrm{goal}$ 
(i.e. is admissible, a necessary condition for A* optimality).

In this paper, the functions $\hat{f}_\mathrm{cost}$ and $\hat{f}_\mathrm{RCOF}$ are given
by an oracle which predicts the value of $f_\mathrm{cost}$ and $f_\mathrm{RCOF}$
obtained at the end of the whole planning pipeline. Notice that in this paper,
contrary to \cite{Brandao2016tro}, $f_\mathrm{cost}$ is the function that will be 
optimized at the full-body trajectory optimization level.
We implement $\hat{f}_\mathrm{cost}$ and $\hat{f}_\mathrm{RCOF}$ as hash tables.
The tables are filled offline, by feeding the whole planning pipeline 
(i.e. trajectory optimization, interpolation, dynamic stabilization) with
uniformly distributed samples of $(s, s', p)$ as shown in Figure \ref{fig:flowchart}.
The discrete optimization problems in (\ref{eq:astarCost}), (\ref{eq:astarHeuristic}) are then
solved for a large number of discretized stances and 
coefficient of friction quantiles and stored in new hash tables for fast access to costs 
and heuristics during search.

\subsection{Full-body trajectory optimization}

The full-body trajectory optimizer takes a footstep plan with $N$ stances and produces a full-body trajectory,
parameterized by $T$ discrete-time waypoints. Waypoints are full-body robot configurations 
$q_t \in \mathbb{R}^D$, $t=1,...,T$, where $D$ is the number of degrees-of-freedom consisting of the 
joints' angle values and the pose of the robot base.
Each stance is associated with 2 full-body postures (at start and midstance) and so $T=2N$.
For convenience we use $s_t$ to refer to the stance associated to $q_t$.

Our optimizer solves the problem
\begin{subequations}
\begin{align}
&\underset{q_1,...,q_T}{\text{minimize}} && f_\mathrm{cost}(q_1,...,q_T) + \alpha f_\mathrm{collision}(q_1,...,q_T) \\
&\text{subject to} \notag \\
&&& f_\mathrm{stance}(q_t,s_t) = 0 \quad \forall_{t \in 1,...,T} \\
&&& f_{xy}(q_t) \in \mathcal{P}_t \quad \forall_{t \in 1,...,T} \\
&&& f_\mathrm{roll}(q_t) = 0 \quad \forall_{t \in 1,...,T} \\
&&& A_t q_t \leq b_t \quad \forall_{t \in 1,...,T} ,
\end{align}
\label{eq:min}
\end{subequations}
where $q_1,...,q_T$ are the optimization variables, $\alpha$ is a penalty constant and:
\begin{itemize}
\item The function $f_\mathrm{cost}$ computes the sum of the squared static torques of all
joints at all waypoints, as implemented in the \textit{trajopt} library \cite{Schulman2014}
\item The function $f_\mathrm{collision}$ is a collision cost as proposed by \cite{Schulman2014}. 
It is the sum of a discrete collision cost computed by the signed distance between 
each link and all other geometries, and a continuous collision cost computed by the signed distance 
between the swept volume of each link with the environment
\item The function $f_\mathrm{stance}(q_t,s_t)$ computes the pose error of all links in contact 
as a $6C$-dimensional vector where C is the number of active contacts in $s_t$. This is 
computed as the translation and axis-angle error between the target link pose 
(given by $s_t$) and the current link pose (given by $q_t$)
\item The function $f_{xy}(q_t)$ computes the (x,y) coordinates of the COM, and $\mathcal{P}_t$ is
the support polygon of $s_t$. The constraint thus enforces approximate static stability.
The support polygon of $s_t$ is computed by the convex hull of the horizontal projection of links in contact
and does not include contacts removed in $s_{t+1}$
\item The function $f_\mathrm{roll}(q_t)$ computes the rotation around the X axis for the waist link,
with respect to the global reference frame. 
This constraint is necessary as ``zero roll'' is an assumption of the subsequent dynamic stabilization 
method (Section \ref{sec:planningStabilization})
\item $A_t, b_t$ enforce joint angle and velocity limits.
\end{itemize}

We solve problem (\ref{eq:min}) using the Sequential Quadratic Programming method
of \cite{Schulman2014} as implemented in the \textit{trajopt} library%
\footnote{URL: http://rll.berkeley.edu/trajopt}.

\subsection{Interpolation and stabilization}
\label{sec:planningStabilization}

To obtain a densely-sampled trajectory for execution on the robot, we interpolate trajectory waypoints 
using Hermite cubic splines with derivatives set to zero for smooth contact transitions. The time between two consecutive waypoints $q_t$ is given by the oracle,
as we describe in Section \ref{sec:planningFootstep}.

Since the obtained trajectory is not dynamically stable, we then apply an FFT-based 
ZMP trajectory compensation scheme \cite{Hashimoto2015}.
The method considers the rigid-body dynamics of the full body and locally adapts COM motion
on the horizontal plane using analytic inverse kinematics to iteratively reduce the error 
between the real and reference ZMP trajectory.
We set the reference ZMP trajectory to the interpolated $f_\mathrm{xy}(q_t)$, which were used in the 
optimization problem (\ref{eq:min}) and are inside the support polygon at each waypoint.
Furthermore, our implementation of the analytic inverse kinematics of the robot WABIAN-2 
assumes zero roll angle of the waist link with respect to the world reference frame. 
We include this constraint in the optimization problem (\ref{eq:min}) for consistency.

\section{RESULTS}
\label{sec:results}

\subsection{Material segmentation results}

To train the CNN we first collected 7,791 annotated images from publicly available semantic-segmentation datasets: 
5,216 from the VOC2010 Context dataset \cite{Mottaghi2014} and  2,575 from the OpenSurfaces dataset \cite{Bell2013}.
We selected all images in the datasets with at least one of the following labels: 
\textit{asphalt, concrete, road, grass, rock, sand, sky, snow, water, carpet, rug, mat, ceramic, tile, cloth, fabric,
marble, metal, paper, tissue, cardboard, wood}.
Due to similarity between some classes at the image and semantic level we joined the labels 
(\textit{asphalt, concrete, road}), (\textit{carpet, rug, mat}), (\textit{ceramic, tile}),
(\textit{cloth, fabric}) and (\textit{paper, tissue, cardboard}). 
The total number of considered classes in the output CNN layer was 14.
\textit{Sky} was only included to avoid classifying it as any of the other materials on outdoor pictures.
\begin{figure*}
\centering
\begin{overpic}[width=0.161\linewidth]{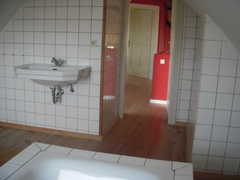} \put (0,5) {\hl{1}} \end{overpic}
\includegraphics[width=0.161\linewidth]{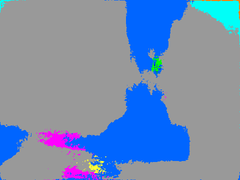}
\includegraphics[width=0.161\linewidth]{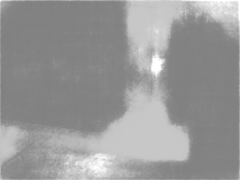}
\begin{overpic}[width=0.161\linewidth]{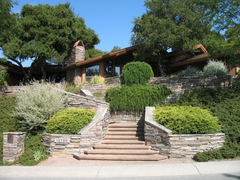} \put (0,5) {\hl{2}} \end{overpic}
\includegraphics[width=0.161\linewidth]{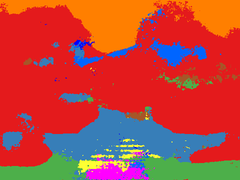}
\includegraphics[width=0.161\linewidth]{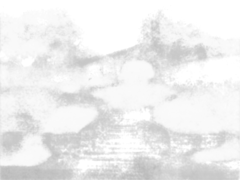}
\begin{overpic}[width=0.161\linewidth]{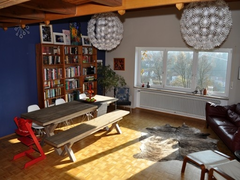} \put (0,5) {\hl{3}} \end{overpic}
\includegraphics[width=0.161\linewidth]{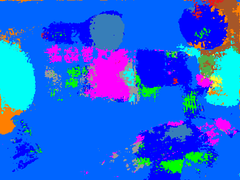}
\includegraphics[width=0.161\linewidth]{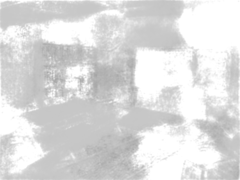}
\begin{overpic}[width=0.161\linewidth]{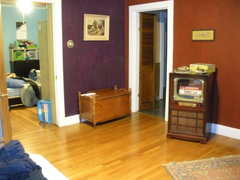} \put (0,5) {\hl{4}} \end{overpic}
\includegraphics[width=0.161\linewidth]{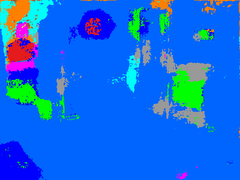}
\includegraphics[width=0.161\linewidth]{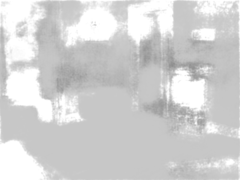}
\begin{overpic}[width=0.161\linewidth]{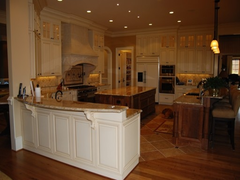} \put (0,5) {\hl{5}} \end{overpic}
\includegraphics[width=0.161\linewidth]{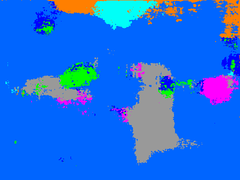}
\includegraphics[width=0.161\linewidth]{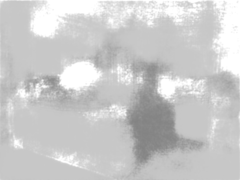}
\begin{overpic}[width=0.161\linewidth]{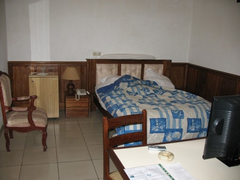} \put (0,5) {\hl{6}} \end{overpic}
\includegraphics[width=0.161\linewidth]{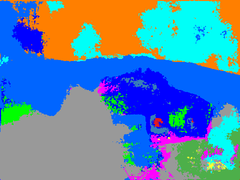}
\includegraphics[width=0.161\linewidth]{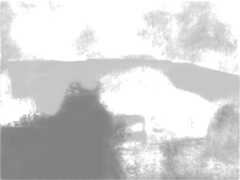}
\begin{overpic}[width=0.161\linewidth]{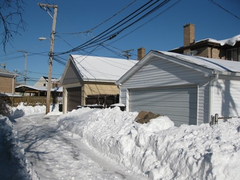} \put (0,5) {\hl{7}} \end{overpic}
\includegraphics[width=0.161\linewidth]{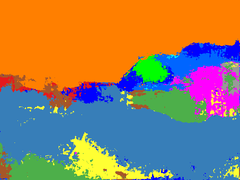}
\includegraphics[width=0.161\linewidth]{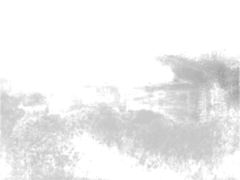}
\begin{overpic}[width=0.161\linewidth]{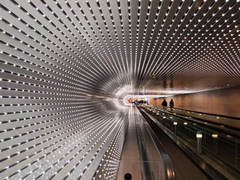} \put (0,5) {\hl{8}} \end{overpic}
\includegraphics[width=0.161\linewidth]{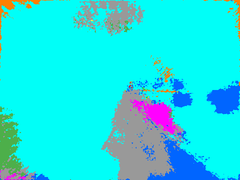}
\includegraphics[width=0.161\linewidth]{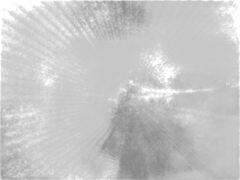}
\begin{overpic}[width=0.161\linewidth]{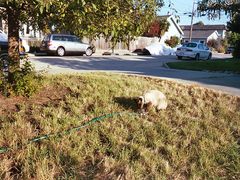} \put (0,5) {\hl{9}} \end{overpic}
\includegraphics[width=0.161\linewidth]{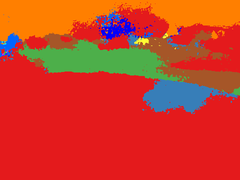}
\includegraphics[width=0.161\linewidth]{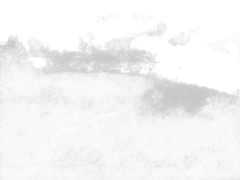}
\begin{overpic}[width=0.161\linewidth]{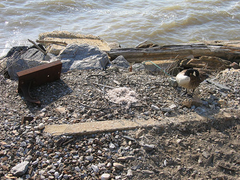} \put (0,5) {\hl{10}} \end{overpic}
\includegraphics[width=0.161\linewidth]{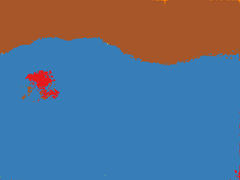}
\includegraphics[width=0.161\linewidth]{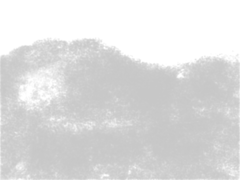}
\begin{overpic}[width=0.161\linewidth]{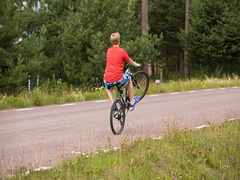} \put (0,5) {\hl{11}} \end{overpic}
\includegraphics[width=0.161\linewidth]{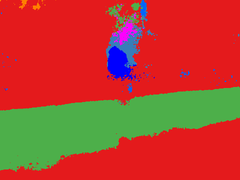}
\includegraphics[width=0.161\linewidth]{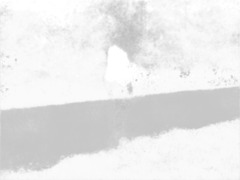}
\begin{overpic}[width=0.161\linewidth]{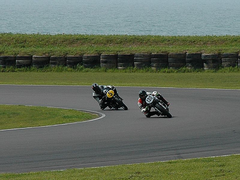} \put (0,5) {\hl{12}} \end{overpic}
\includegraphics[width=0.161\linewidth]{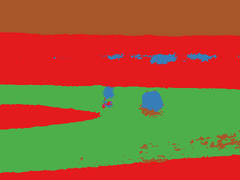}
\includegraphics[width=0.161\linewidth]{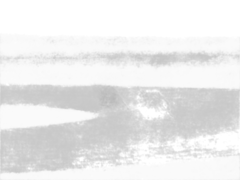}
\begin{overpic}[width=0.161\linewidth]{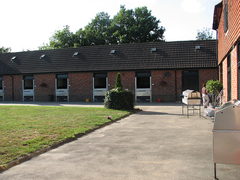} \put (0,5) {\hl{13}} \end{overpic}
\includegraphics[width=0.161\linewidth]{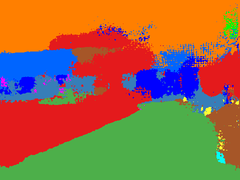}
\includegraphics[width=0.161\linewidth]{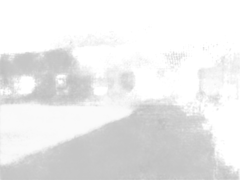}
\begin{overpic}[width=0.161\linewidth]{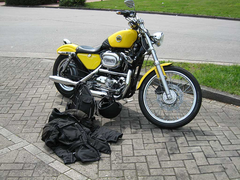} \put (0,5) {\hl{14}} \end{overpic}
\includegraphics[width=0.161\linewidth]{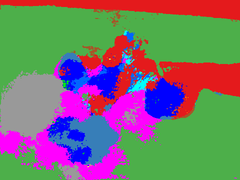}
\includegraphics[width=0.161\linewidth]{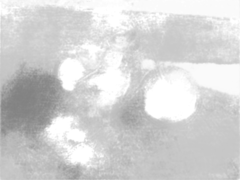}
\includegraphics[width=0.60\linewidth]{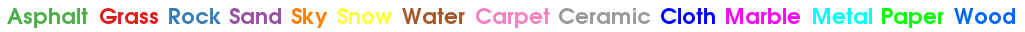}
\caption{Example images from the test set, respective material segmentation 
	(highest probability labels) and coefficient of friction quantiles $Q_{1-0.95}$
	of equation (\ref{eq:friction}). Darker shades of gray correspond to higher friction, 
	such that white is $\mu=0$ and black $\mu=1$. }
\label{fig:materialfriction}
\end{figure*}

We used stochastic gradient descent with 0.1 learning rate and 0.9 momentum as in the
original SegNet publication \cite{Badrinarayanan2015}, and trained the network on an Amazon
Elastic Cloud node with a 4GB NVIDIA GPU. We ran a total of 90,000 iterations with a mini-batch 
size of 5 (maximum allowed by the GPU). Training was done on 60\% of the images, while the other
40\% were used as the test set.

We obtained a global classification accuracy of 0.7929 and class-average accuracy of 0.4776 on the test set.
See Figure \ref{fig:materialfriction} for examples of the (highest probability) material predictions given by the CNN
on the test set.
The global accuracy is comparable to state-of-the-art performance in semantic segmentation
(e.g. \cite{Badrinarayanan2015,Bell2015}), and the class-average accuracy is slightly below
state-of-the-art (which is around 0.60 \cite{Badrinarayanan2015}). We believe one important way 
to improve classification accuracy is to improve the dataset itself since, for instance, 
there is moderate visual similarity between some of the materials such as marble and ceramic, 
and some materials are lowly sampled (e.g. the lowest sampled materials are snow and sand, 
present in 173 and 46 images respectively).

The material segmentation results in Figure \ref{fig:materialfriction} show an overall good accuracy
of the CNN, particularly on wood, grass and sky labels. The figure also shows typical misclassifications 
such as white walls recognized as sky or metal (picture 6), hard snow as rock (picture 7), and
some overlap between asphalt/road, ceramic and marble. These are arguably understandable
since material labels themselves semantically overlap. However, our approach to the visual friction 
estimation problem is such that if there is uncertainty in the material label, then this uncertainty 
can be used to weight the friction of the surface through material and friction probability 
distributions (Section \ref{sec:vision}).

\subsection{Friction prediction results}

\begin{figure}
\centering
\adjincludegraphics[width=0.80\linewidth,trim={0 0 0 {.4\height}},clip]{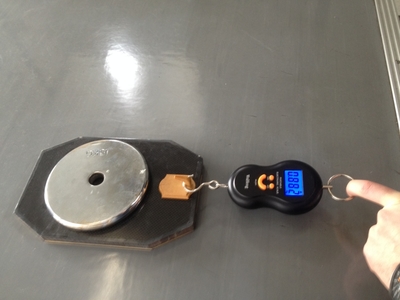}
\caption{We estimate the coefficient of friction of a material by several measurements of the
	maximum friction force on the robot foot, loaded with a 1.5kg mass.}
\label{fig:frictionmeasurement}
\end{figure}

We empirically measured the coefficient of friction associated with each material label using a force gauge
and the robot foot loaded with a 1.5kg mass. 
The foot is rigid and its sole is covered with a high stiffness soft material for shock absorption and an anti-slippage sheet.
We checked whether surfaces were horizontal with a level,
then placed the foot and measured maximum friction force with the force gauge.
See Figure \ref{fig:frictionmeasurement} for an illustration of the procedure.
We took 5 friction measurements on each surface, and used at least 3 surfaces of each material.
We fitted a normal distribution to the measurements, obtaining separate parameters
$\mu_i$ and $\sigma_i^2$ for each material, where $\mu_i$ is the mean friction of material $i$, 
and $\sigma_i^2$ the variance.
The materials \textit{sand, snow, water, cloth, paper} were an exception, and since our robot
is currently not capable of walking on them (i.e. fall or damage risk is too high) we directly set them to 
$\mu_i=0$, $\sigma_i^2=0$. We similarly set \textit{sky}'s friction to zero as well.
See Table \ref{table:cof} for the parameters of the friction p.d.f. of each material.
\begin{table}
	\begin{center} 
		\caption{ Normal distribution parameters of each material's coefficient of friction, 
			measured manually on the robot foot } 
		\label{table:cof} 
		\begin{centering} 
			\begin{tabular}{cccccc} 
				\toprule 
				Material	& $\mu_i$  & $\sigma_i$ \\
				\midrule
				Asphalt		& 0.74				& 0.12 \\ 
				Grass		& 0.53				& 0.10 \\ 
				Rock		& 0.80				& 0.08 \\ 
				Carpet		& 0.82				& 0.02 \\ 
				Ceramic		& 0.97				& 0.05 \\ 
				Marble		& 0.83				& 0.15 \\ 
				Metal		& 0.80				& 0.15 \\ 
				Wood		& 0.88				& 0.12 \\
				Sand, Sky, Snow, Water, Cloth, Paper* & 0 & 0 \\
				\bottomrule 
			\end{tabular}
		\end{centering}
	\end{center}
	Note: materials marked with a * are assumed to be untraversable by our robot and so set to zero without measurements.
\end{table}

In Figure \ref{fig:materialfriction} we show the test-set's highest probability material predictions 
along with the $(1-\eta)$-quantile of the coefficient of friction which is used in equation 
(\ref{eq:quantile}). We set a typical value of $\eta=0.95$.
The friction images are darker where friction is higher ($\mu=1$ would be black). 
Note that ceramic-like surfaces have high predicted friction (pictures 1, 5, 6); 
beds and jackets have very low friction (pictures 3, 4, 14); 
grass patches have lower friction than roads (pictures 2, 9, 11, 12, 13); 
and that water is mostly white - zero friction - (pictures 10, 12).

The figure also shows the advantage of using the whole probability distribution of materials
(instead of using the highest probability material) to estimate friction. 
For example in picture 14, the jacket on the ground is classified as cloth and rock depending on the region, 
but friction is low on most of the object's area since the \textit{cloth} label still has high probability.

\subsection{Planning results}

\begin{figure}
\centering
\begin{overpic}[width=0.32\linewidth]{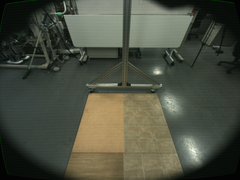} 
	\put (38,20) {\textcolor{white}{x}} \end{overpic}
\includegraphics[width=0.32\linewidth]{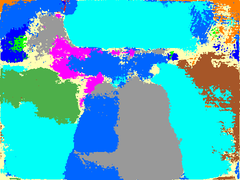}
\includegraphics[width=0.32\linewidth]{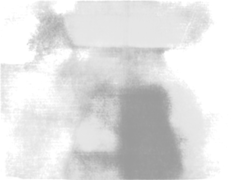} 
\begin{overpic}[width=0.32\linewidth]{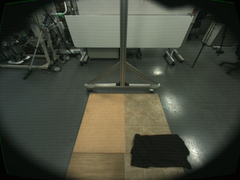} 
	\put (38,20) {\textcolor{white}{x}} \end{overpic}
\includegraphics[width=0.32\linewidth]{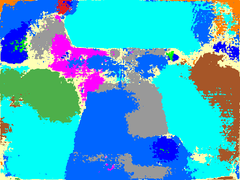}
\includegraphics[width=0.32\linewidth]{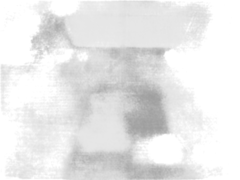}
\includegraphics[trim={0 0 0 -3mm},width=0.70\linewidth]{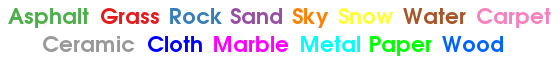}
\caption{Left: the view from the robot's camera of the mock-up scenario built in the laboratory. 
	The locomotion target is one meter ahead, marked with a white ``x''.
	Middle: material segmentation.
	Right: coefficient of friction quantiles $Q_{1-0.95}$ of equation (\ref{eq:friction}). 
	Darker shades of gray correspond to higher friction, such that white is $\mu=0$ and black $\mu=1$.}
\label{fig:planning0}
\end{figure}

We prepared a mock-up scenario in the laboratory which demonstrates the capabilities of our planner.
The scenario consists of a floor with two areas of different materials. One is made of wood 
($\mu=0.84$) and the other is a high-friction flooring resembling ceramic tiles both in 
appearance and coefficient of friction ($\mu=1.00$). 
The perception-planning algorithms were run on this scenario, and then a piece of cloth (T-shirt) was laid flat
on one of the surfaces to provoke changes in friction and force a different plan.
See Figure \ref{fig:planning0} for the scenario, segmentation and friction as seen from the robot's camera
at the initial condition.
Once again, the figures show the advantage of using the full probability distribution of materials
given by the CNN. While \textit{cloth} is the highest-ranking material only in part of the object
region, friction is low on a larger region which is highly consistent with object borders.

The robot starts in double-support, with one foot on each surface. The goal stance is one meter ahead, 
also with a foot on each surface.
After the robot is placed at the initial state,
the perception and planning algorithms run without any human input except the push of a button to execute
the planned full-body trajectory open-loop.
Trajectory optimization parameters are the collision penalty weight $\alpha$
of equation (\ref{eq:min}), which is set to 50, and the distance at which the collision
penalty starts being applied (for all links except those in contact), which we set to 2.5cm.
The obtained full-body trajectory is tracked by position control at the joint level.

For these experiments we used the human-sized humanoid robot WABIAN-2 \cite{Ogura2006} customized with a 
Carnegie Robotics Multisense SL sensor-head.
The perception pipeline predicts pixel-wise material label distributions and pixel-wise friction
using SegNet \cite{Badrinarayanan2015} and equation (\ref{eq:friction}), and combines them with the
stereo depth maps computed onboard by the Multisense.
It produces friction-annotated point clouds at 2Hz.
For collision checking, the point cloud is converted into a mesh using the fast surface
reconstruction algorithm of \cite{Holz2013} as implemented in PCL \cite{PCL}.
All perception and planning computation ran on an external PC with network connection
to the robot's onboard PC, and we used ROS \cite{ROS} for communication.

\begin{figure*}
\centering
\includegraphics[width=0.191\linewidth]{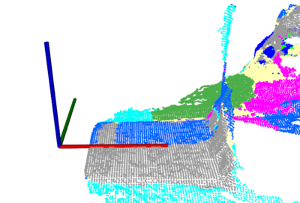}
\includegraphics[width=0.191\linewidth]{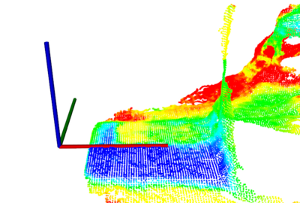}
\includegraphics[width=0.191\linewidth]{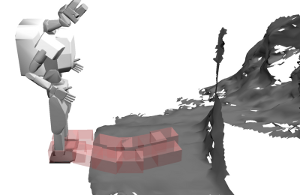}
\includegraphics[width=0.191\linewidth]{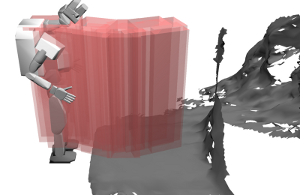}
\includegraphics[width=0.191\linewidth]{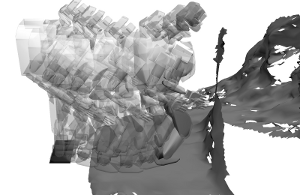}
\adjincludegraphics[width=0.115\linewidth,trim={{.2\width} 0 {.2\width} 0},clip]{planning-experiments/videos-img/00082/00082-11.jpg}
\adjincludegraphics[width=0.115\linewidth,trim={{.2\width} 0 {.2\width} 0},clip]{planning-experiments/videos-img/00082/00082-21.jpg}
\adjincludegraphics[width=0.115\linewidth,trim={{.2\width} 0 {.2\width} 0},clip]{planning-experiments/videos-img/00082/00082-31.jpg}
\adjincludegraphics[width=0.115\linewidth,trim={{.2\width} 0 {.2\width} 0},clip]{planning-experiments/videos-img/00082/00082-41.jpg}
\adjincludegraphics[width=0.115\linewidth,trim={{.2\width} 0 {.2\width} 0},clip]{planning-experiments/videos-img/00082/00082-51.jpg}
\adjincludegraphics[width=0.115\linewidth,trim={{.2\width} 0 {.2\width} 0},clip]{planning-experiments/videos-img/00082/00082-61.jpg}
\adjincludegraphics[width=0.115\linewidth,trim={{.2\width} 0 {.2\width} 0},clip]{planning-experiments/videos-img/00082/00082-71.jpg}
\adjincludegraphics[width=0.115\linewidth,trim={{.2\width} 0 {.2\width} 0},clip]{planning-experiments/videos-img/00082/00082-80.jpg}
\includegraphics[width=0.191\linewidth]{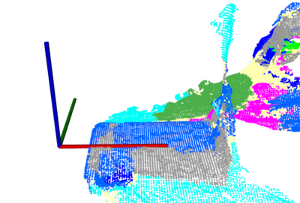}
\includegraphics[width=0.191\linewidth]{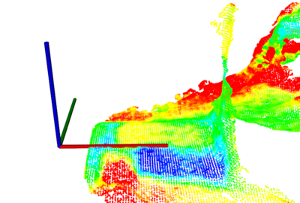}
\includegraphics[width=0.191\linewidth]{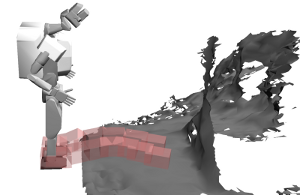}
\includegraphics[width=0.191\linewidth]{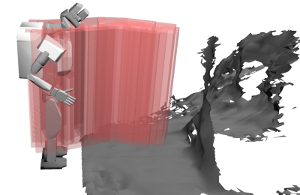}
\includegraphics[width=0.191\linewidth]{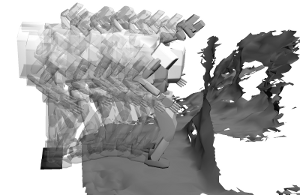}
\adjincludegraphics[width=0.115\linewidth,trim={{.2\width} 0 {.2\width} 0},clip]{planning-experiments/videos-img/00083/00083-21.jpg}
\adjincludegraphics[width=0.115\linewidth,trim={{.2\width} 0 {.2\width} 0},clip]{planning-experiments/videos-img/00083/00083-31.jpg}
\adjincludegraphics[width=0.115\linewidth,trim={{.2\width} 0 {.2\width} 0},clip]{planning-experiments/videos-img/00083/00083-41.jpg}
\adjincludegraphics[width=0.115\linewidth,trim={{.2\width} 0 {.2\width} 0},clip]{planning-experiments/videos-img/00083/00083-51.jpg}
\adjincludegraphics[width=0.115\linewidth,trim={{.2\width} 0 {.2\width} 0},clip]{planning-experiments/videos-img/00083/00083-61.jpg}
\adjincludegraphics[width=0.115\linewidth,trim={{.2\width} 0 {.2\width} 0},clip]{planning-experiments/videos-img/00083/00083-71.jpg}
\adjincludegraphics[width=0.115\linewidth,trim={{.2\width} 0 {.2\width} 0},clip]{planning-experiments/videos-img/00083/00083-81.jpg}
\adjincludegraphics[width=0.115\linewidth,trim={{.2\width} 0 {.2\width} 0},clip]{planning-experiments/videos-img/00083/00083-90.jpg}
\caption{Perception and planning experiments with two surfaces (wood and ceramic)
	on the first two rows and three surfaces (wood, ceramic and cloth) on
	the last two rows. We show the material segmentation (same colour codes
	as in Figure \ref{fig:planning0}), friction (cold colours are high friction, 
	warm are low), footstep plan, collision bounding boxes, full-body plan
	and finally the walking sequence on the real robot.}
\label{fig:planning1}
\end{figure*}

We show the results of the perception-planning experiments in Figure \ref{fig:planning1}.
From left to right, we show the material and friction point clouds, the footstep plan, the collision-checking
bounding boxes used by the footstep planner and the final planned full-body trajectory after optimization and stabilization.
In the first situation there are only wood and ceramic surfaces, but the predicted lower bound
of friction of the wood surface is lower than that of the tiles ($Q_{1-0.95}=$0.1 vs 0.4).
The footstep planner returns a sequence of stances that reduces the amount of times wood is
stepped on. This behavior comes naturally from the \textit{extended footstep planning} approach \cite{Brandao2016tro}, 
since walking on low friction ground requires higher stance times (slower motion)
and thus more energy cost.
Furthermore, note that the trajectory optimization uses all degrees-of-freedom to satisfy the constraints
(e.g. trunk roll use is clear in the image sequence, important mainly for the stability constraints), and
that the knees are relatively stretched in order to reduce torque consumption but still satisfy stability constraints. Also note that swing leg clearance happens automatically due to the use of collision costs.

In the second situation we laid a flat piece of cloth on a ceramic spot used by the previous trajectory. 
The cloth was correctly classified and its friction was practically zero.
The footstep planner returned a trajectory around the cloth and on the wood surface, which led
to a slightly longer time and energy cost of the full-body trajectory (63 vs 60 seconds, 5\% longer than on the first situation).
Note that while the times are long they correspond to 25 stances because of step length limits,
and thus the average time per stance is approximately 2.5 seconds.

Footstep planning took approximately 20 seconds in the first situation and 10 in the second. 
The reason for the difference is clear from the scenario: while in the first situation stances on 
both surfaces are expanded by A* in order to guarantee optimality, in the second situation 
no stances are expanded on the surface with cloth since friction zero has infinite cost.
Full-body trajectory optimization took approximately 40 seconds and dynamic stabilization 2 seconds.
Note that these are for 25-stance, 60 second trajectories, and therefore they should be
considerably faster in case planning is done one or two steps at a time.

\section{CONCLUSIONS AND DISCUSSION}
\label{sec:conclusions}

In this paper we proposed a complete solution to the problem of biped robot locomotion
on slippery terrain. We developed both a visual friction estimation algorithm and an 
objective-consistent hierarchical planning method which considers trajectory costs, 
collision, stability and friction.

We empirically showed that friction estimates in our algorithm are more consistent with 
object/material borders than the highest-probability material label segmentation, which 
shows a good integration of segmentation uncertainty into friction estimation.
We also showed that the algorithms work for varied terrain and are applicable to planning 
on a real robot. The algorithms are relevant since not only obstacles but also different 
terrain types abound in the real world, and locomotion choices should take them into 
account - whether for safety or energetic considerations.

Regarding the perception problem, we opted to decouple it into (broad sense) material 
segmentation and per-material friction distributions. Even though we obtained the 
material friction distributions manually, these could also be learned over time with 
locomotion experience.
Alternatively, friction could also be learned from images directly by end-to-end training,
for example by initialization of a CNN with the parameters obtained with our architecture.

For the context of this paper all surfaces were dry. Wet surfaces could also be included,
although from our experience they should be treated as separate material labels 
(e.g. ``dry metal'' and ``wet metal'') so that the distribution $\mu|l_i$ does not become bimodal.
Thus, one important detail in this work is the notion of material, which should be taken 
in a broad sense, as a visually distinguishable terrain class.

Importantly, one problem with the proposed perception and planning approach is that 
wrong material classifications can lead to there being no solution to the footstep 
planning problem. An example of such a situation is when a material the robot cannot walk on, 
such as water in our case, is mistakenly given very high confidence.
Our view is that the solution could be semi-supervision where a teleoperator can correct a 
segmented region's material label. However, we believe that humans should not directly 
annotate COF, since despite their relative ability to adapt gait to slippery ground humans 
have difficulties in estimating coefficient of friction values \cite{Lesch2008,Brandao2016friction}.

Finally, full-body trajectories in this paper were interpolated after trajectory optimization 
at waypoints. While we did this for implementation simplicity, one possible direction of 
improvement could be to use the spline representation directly in the optimization problem, 
using constraints at collocation points.


\addtolength{\textheight}{-10cm}   


\bibliographystyle{IEEEtran}

\end{document}